\documentclass[a4paper,13pt]{article}

\usepackage{quoting}
\usepackage[utf8]{inputenc} 
\usepackage[T1]{fontenc}    
\usepackage{hyperref}       
\usepackage{url}            
\usepackage{booktabs}       
\usepackage{amsfonts}       
\usepackage{nicefrac}       
\usepackage{microtype}      
\usepackage{subfigure}
\usepackage{amsmath}
\usepackage{apalike}
\usepackage{color}
\usepackage{epsfig}

\newcommand\shrink[1]{}

\usepackage[a4paper, total={5.5in, 8in}]{geometry}

\title{Human-Level Intelligence or Animal-Like Abilities?\thanks{Submitted to Communications of the ACM. \newline
A talk on this article can be found at \url{https://www.youtube.com/watch?v=UTzCwCic-Do}}}

\author{
 Adnan Darwiche \\
 Computer Science Department \\
 University of California, Los Angeles \\
  \texttt{darwiche@cs.ucla.edu} \\
}

\date{July 8, 2017}

\begin{document}

\maketitle

\begin{quote}
The vision systems of the eagle and the snake outperform everything that we can make in the laboratory, but snakes and eagles cannot build an eyeglass or a telescope or a microscope.

\hspace{7cm} Judea Pearl\footnote{Lecture on {\em The Mathematics of Causal Inference, with Reflections on Machine Learning and the Logic of Science,} 
\url{https://www.youtube.com/watch?v=zHjdd--W6o4}. 
}
\end{quote}

\section{Introduction}

The recent success of neural networks in applications such as speech recognition, vision and autonomous navigation has led to great excitement by 
members of the artificial intelligence (AI) community and the general public at large. 
Over a relatively short period, by the science clock, we managed to automate some tasks that have defied us for decades and using one of the more classical 
techniques coming out of artificial intelligence  research. 

The triumph over these achievements has led some to describe the automation of these tasks as having reached human level intelligence.
This perception, originally hinted at in academic circles, has been gaining momentum more broadly and is leading to some implications. For example,
a trend is emerging in which machine learning research is being streamlined into neural network research, under its newly acquired label ``deep learning.''
This perception has also caused some to question the wisdom of continuing to invest in other machine learning
approaches, or even mainstream areas of artificial intelligence, such as knowledge representation, symbolic reasoning and planning. Some coverage
of AI in public arenas, particularly comments made by some visible figures, has led to mixing this excitement with fear of what AI may be
bringing us in the future (i.e., doomsday scenarios).\footnote{Stephen Hawking said, ``The development of full artificial intelligence could spell the end of the human race''
and Elon Musk said that AI is ``potentially more dangerous than nukes.''}

This turn of events in the history of AI has created a dilemma for researchers in the broader AI community. 
On the one hand, one cannot but be impressed with, and enjoy, what we have been able to accomplish with neural networks. On the other hand, mainstream scientific intuition
stands in the way of accepting that a method, which does not require any explicit modeling or sophisticated reasoning, 
can be sufficient for reproducing human level intelligence. This dilemma is further amplified by the observation that recent developments did not culminate in a clearly characterized 
and profound scientific discovery---such as a new theory of the mind---that would normally mandate massive updates to the AI curricula.
Scholars from outside AI and computer science often sense this dilemma as they complain that they are not receiving an intellectually satisfying
answer to the question of ``What just happened in AI?''

The answer to this dilemma lies in a careful assessment of what we managed to achieve with deep learning,
and in identifying and appreciating the key scientific outcomes of recent developments in this area of research. This has been unfortunately lacking to
a great extent. My aim here is to trigger such a discussion, encouraged by the positive and curious feedback I have been receiving on the thoughts expressed in this article.

\begin{figure}[t]
  \centering
  \includegraphics[height=0.2\textheight]{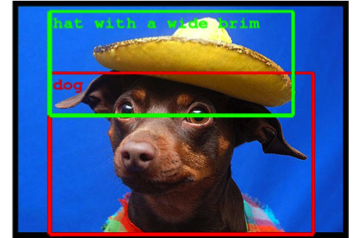} \qquad
  \includegraphics[height=0.2\textheight]{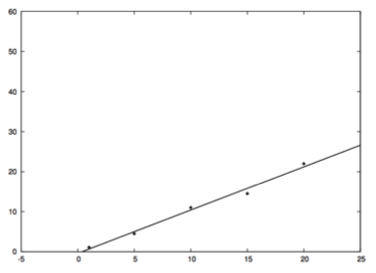}
  \caption{\small Left: Object recognition \& localization (ImageNet). Right: Fitting a function to data.}
  \label{fig:function}
\end{figure}

\section{Background}

To lay the grounds for this discussion, I will first mark two distinct approaches for tackling problems that have been of interest to AI. I will call the first approach {\em model-based}
and the second {\em function-based.} 
Consider the object recognition and localization task in Figure~\ref{fig:function}. To solve this recognition task, the model-based approach requires one to represent knowledge about
dogs and hats, among other things, and involves reasoning with such knowledge. The main tools of this approach today are logic and probability (mathematical modeling more generally) 
and it can be thought of as the represent-and-reason approach originally envisioned by the founders of AI. It is also the approach normally expected, at some level, by the informed member 
of the scientific community. The function-based approach on the other hand formulates this as a function-fitting problem, with the function inputs coming directly from the image pixels
and its outputs corresponding to the high-level recognitions we are seeking (the function must have a form that can be evaluated efficiently). 
The main tool of this approach today is the neural network.
Many college students have exercised a version of this approach in a physics or a chemistry lab, where they fit simple functions to data collected from various experiments; 
see Figure~\ref{fig:function}. The main difference here is that we are now employing functions that have multiple inputs and outputs; the structure of these functions can be arbitrarily complex; 
and the problems we are tackling are ones that we tend to associate with cognition, as opposed to, say, estimating the relationship between volume and pressure in an engineered 
system.\footnote{Some refer to this as the {\em curve-fitting} approach. While this term highlights the efficient evaluation of a function (the ``curve'') and captures the spirit of the function-based approach, it underplays
the complex and rich structure of functions encoded by today's (deep) neural networks.}

The main observation in AI during the last few years is that the function-based approach can be quite effective at certain AI tasks,
more so than the model-based approach. 
This has surprised not only mainstream AI researchers, who mainly 
practice the model-based approach, but also machine learning researchers who practice various approaches, of which the function-based approach is just one.\footnote{Machine learning 
includes the function-based approach but has a wide enough span that overlaps with the model-based approach. For example, one can learn the parameters or even structure of
a model, but that may still require complex reasoning to obtain answers from the model (i.e., efficient evaluation is neither required nor guaranteed).}
This has had many implications, some positive and some giving grounds for concern.

On the positive side is the increasing number of tasks and applications that are now within reach, using a tool that can be very familiar to someone with only a broad 
engineering background---particularly, one who is accustomed to estimating functions and using them to make predictions. What is of concern, however, is the current imbalance
between exploiting, enjoying and cheering this tool on the one hand, and thinking about it on the other hand. This thinking is not only important for realizing the full potential
of this tool, but also for scientifically characterizing and understanding its potential reach. 
The lack of such characterization, or at least the attempt to seek it aggressively enough, is a culprit of current misconceptions about AI progress and where that may lead us in the future.

\section{What Just Happened in AI?}

In my own quest to fully appreciate the progress enabled by deep learning, I came to the conclusion that recent developments tell us more about
the {\em problems tackled} by deep learning than they tell us about neural networks per se. These networks are parametrized 
functions that are expressive enough to capture any relationship between inputs and outputs, and have a form that can be evaluated efficiently. 
This we have known for decades and is detailed in various AI textbooks. What caused the current turn of events then? 

To shed some light into this question, let me state again what we have discovered recently. That is, some seemingly complex  abilities that are typically
associated with cognition can be captured and reproduced to a reasonable extent by simply fitting functions to data, i.e., without the need for an explicit 
modeling of the environment or symbolically reasoning about it. While this is a remarkable finding, it highlights {\em problems} and {\em thresholds} 
more than it highlights {\em technology.} Let me explain. Every behavior, intelligent or not, can be captured by a function 
that maps inputs (environmental sensing) to outputs (thoughts or actions). The two key questions though are the following.
Are these functions simple enough to admit a compact representation that allows mapping inputs to outputs efficiently as in neural networks (i.e., without the need for reasoning)?
And if the answer is yes, do we currently posses the ability to estimate these functions from input-output pairs (i.e., data)? 

What has happened in AI recently are three developments that bear directly on these questions. The first development is our improved ability to fit functions to data,
which has been enabled by (a) the availability of massive amounts of data; (b) the increased computational power we now have at our hands; and
(c) the increasingly more sophisticated statistical techniques for fitting functions.
The second development is that we have identified a class of practical applications which correspond to functions that, we now know, are simple enough to allow compact
representations that can be evaluated efficiently (again, without the need for reasoning), and whose estimation is within the reach of current thresholds for gathering data, 
computational speed and estimation techniques. This includes identifying and localizing objects in some classes of images and certain tasks that pertain to natural language and speech.
The third development, which goes largely unnoticed, is that we changed our measures for success in ways that
reduced the technical challenges considerably, at least as entertained by early AI researchers,
while maintaining our ability to capitalize on the obtained results commercially (I will discuss this last point in more detail later).

Interestingly, none of the above amounts to a major technical breakthrough or milestone in AI per se, such as the establishment of probability as a foundation of commonsense reasoning
in the late 1980s, or the introduction of neural networks more than fifty years ago.\footnote{It is worth mentioning here that research on neural networks has gone through many turns 
since their early traces in the 1940s. For example, early results on the limitations of the perceptron (a shallow neural network) almost stopped research 
on this subject in the 1960s, only to be resumed more actively in the 1970s when the first multi-layer (deep) networks were put to use.} Yet, the combination of these
factors has had a profound impact on real-world applications and the successful deployment of various AI techniques (in conjunction with neural networks)
that have been in the works for a very long time.

\section{I Beg to Differ}

I shared the above remarks in various contexts during the course of preparing this article. The audiences ranged from AI, computer science, to 
law and public policy researchers with an interest in AI. What I found striking is the great interest in this discussion and the comfort, if not
general agreement, with the remarks I made. I did get a few ``I beg to differ'' responses though. They all centered around recent advancements relating 
to optimizing functions, which are key to the successful training of neural networks (e.g., results on stochastic gradient descent, dropouts, etc).
The objections stemmed from not having named these as breakthroughs or milestones (in AI). My answer: These all fall under the enabler 
I listed above, ``more sophisticated statistical techniques for fitting functions.'' 
Follow up question: Does it matter that they are statistical techniques as opposed to classical AI techniques? 
Answer: It does not matter as far as acknowledging and appreciating scientific inquiry and progress, but it does matter as far as explaining what just happened and, 
more importantly, forecasting what may happen next. 

Consider an educated individual
sitting next to you, the AI researcher, on a plane (I get that a lot). They figure out you do AI and then ask: What are the developments that enabled the 
current progress in AI? You tell them the function-based story and lay out the three enablers. They will likely be impressed and also intellectually satisfied. 
However, if the answer was ``we just discovered a new theory of the mind,'' 
you will likely not be surprised if they also end up worrying about a Skynet coming
soon to mess up our lives. Public perceptions about AI progress and its future are very important. The current misperceptions and associated fears 
are being nurtured by the absence of scientific, precise and bold perspectives on what just happened, therefore, leaving much to the imagination.

This is not to suggest that {\em only} a new theory of the mind, or an advance of that scale, would justify some of the current and legitimate concerns
surrounding AI.
In fact, even limited AI technologies can entice autonomous systems that may pose all kind of risks. These concerns, however, are not new to our 
society---e.g., safety considerations when auto pilots were introduced into the aerospace industry and job considerations when  ATMs were introduced 
into the banking industry. The headline here should therefore be ``automation'' more so than ``AI,'' as the latter is just a technology that happened to improve and speed up automation.
As such, to address these concerns, the focus should be shifted towards policy and regulatory considerations for dealing with the new level of automation 
that our society is embarking on, instead of fearing AI.\footnote{Mike Wellman noted to me that to the extent the success of deep learning has been surprising 
in many ways, it should remind us that advances in AI can come suddenly and we should be prepared.}

\section{Moving the Goalposts}

Let me now address the third reason for the current turn of events, which is the change in how we measure success. This particular reason is quite substantial
yet goes largely unnoticed, especially by younger researchers.

Consider machine translation for example,
which received significant attention in the early days of AI. The represent-and-reason approach is considered to have failed on this task, with machine learning approaches
being the state of the art now (going beyond function-based approaches). In the early days of AI, success was measured by how far a system accuracy was from \(100\%\),
with intelligence being a main driving application  (a failure to translate correctly can potentially lead to a political crisis). 
The tests that translation systems were subjected to compared their performance to human translation abilities (e.g., translate a sentence from English to Russian then back to English).
Today, the almost sole application of machine translation is to the web. Here, success is effectively measured in terms of how far a system accuracy is from \(0\%\). 
If I am looking at a page written in French, a language that I don't speak, I am happy with any translation that gives me a sense of what the page is saying. 
In fact, the machine translation community rightfully calls this {\em gist translation.} It can work impressively well on prototypical sentences that appear often in the data, but can
fail badly on novel text. This is still very valuable, yet it corresponds to a task that is significantly different from what 
was tackled by early AI researchers. Current translation systems will fail miserably if subjected to the tests adopted by early translation researchers.
Moreover, these systems will not be suitable, or considered successful, if integrated within a robot that is meant to emulate human behavior or abilities.

The ``Allen AI Science Challenge'' introduced recently puts this observation in further perspective. The task here is to comprehend a paragraph that states a science problem
at the middle school level and then answer a multiple choice question. There are numerous reports on the success of deep learning when
applied to text in various contexts. But can these systems pass the Allen AI Science test at the level of a competent middle schooler?

Similar observations can be made about speech recognition systems. The main application of these systems today is in user interfaces, such as automated technical support  
and the commanding of software systems (e.g., phone and navigation systems in vehicles). Have you tried these? They fail and often (just try to say something
that is not very prototypical or try not to hide your accent if you have one). But when these systems fail, they send you back to a human operator or force you to command the 
software using classical means. Is this still practically useful? Yes. Does it measure up to the expectations of AI as a field? No. 

Moving to vision applications, reports are out there on how making simple changes to images may sometimes break the ability of neural networks to recognize
objects correctly. Some transformations or deformations to objects in images, which maintain the human ability to recognize them, can also break the ability of 
networks to recognize them. While this does not measure up  to the expectations of early AI researchers, or even contemporary vision researchers,
we still manage to benefit from these technologies in applications such as recognizing faces during auto focus in smart cameras (people don't normally deform their 
faces, but if they do, bad luck, an unfocused image); looking up images that contain cats in online search (it is ok if you end up getting a dog instead), 
and localizing surrounding vehicles in an image taken by the camera of a self driving car (the vulnerability of these systems to mistakes remains a
controversial one in both its scope and how to deal with it at the policy or regulatory levels). 

The significance of these observations stems from their bearing on our ability to forecast the future and our decisions on what research to invest in. 
In particular, does the success in addressing these selected tasks, which are driven by circumscribed commercial applications, justify the worry about doomsday scenarios? 
Does this kind of success justify claims that we have figured out a new way to formalize commonsense reasoning? 
Does it justify claims that we can now comprehend language or speech or do vision at the levels that humans do? My own answer is no.
Does it justify this current imbalance of attitudes towards various machine learning and AI approaches? If you work for a company that has an interest in one
of these applications, then the answer is perhaps, and justifiably, yes. However, if you are concerned with scientific inquiry more broadly, then the answer is hopefully no.

In summary, what just happened in AI is nothing close to a breakthrough that justifies worrying about doomsday scenarios. 
What just happened is the successful employment of AI technology in a selected class of commercial applications, 
aided greatly by developments in sister fields, and by performance requirements that are looser than originally sought.
Put another way --- and in response to titles I see today, like ``AI has arrived,'' ``I didn't see AI was coming'' --- what really came are
numerous applications that can benefit from improved AI techniques, that still fall short of AI ambitions, but are good enough to be 
capitalized on commercially. This by itself is positive, up until we start confusing it with something else.

Let me close this section by stressing two points. 
The first is to reemphasize an earlier observation that while current AI technology is still quite limited, 
the impact it may have on automation and, hence, society may be substantial (e.g., jobs and safety).
This in turn calls for profound treatments at the technological,\footnote{The recent White House initiative, ``Preparing for the Future of AI,''  particularly raised awareness 
about the {\em safety} of AI systems, a term that was mentioned dozens of times in the final report of this initiative. We now hear about efforts such as ``safe reinforcement learning''
and ``verified AI systems,'' which highlight how safety considerations are starting to impact AI technology.} policy and regulatory 
levels.\footnote{Eric Horvitz brought up the idea of subjecting certain AI systems to {\em trials} just as we do to approve drugs. 
I think that proper {\em labeling} of certain AI systems should also be considered, again as we do with drugs. For example, some have suggested that the term ``self driving
car'' is perhaps responsible for the misuse of this technology by some drivers, who are expecting more from this technology than is currently warranted.}
The second point I wish to stress is that ``moving the goalposts'' has been an enabling and positive development that we need to be accutely aware of and better understand its implications.
A key finding/implication here is that some cognitive tasks can be emulated to a reasonable extent without the need to understand 
or formalize these cognitive tasks as originally believed and sought (as in some speech and vision applications). 
That is, we succeeded in these applications by having circumvented certain technical challenges instead of having solved them directly. This observation is not meant
to discount current success, but to highlight its nature and lay the grounds for the following question:
How far can we go with this direction? I will revisit this issue later.

\section{Human-Level or Animal-Level?}

Let me now get to the thoughts that triggered the title of this article in the first place. 
I believe that attributing human level intelligence to the tasks currently conquered by neural networks is questionable, 
as these tasks barely rise to the level of abilities possessed by many animals. Judea Pearl already cited eagles and snakes as having vision systems that 
surpass what we can build today. Cats have navigation abilities that are far more superior to any of the ones possessed by existing,
automatous navigation systems including self driving cars. Dogs can recognize and react to human speech and African grey parrots can generate sounds that mimic human speech
to impressive levels. Yet none of these animals posses the cognitive abilities and intelligence typically attributed to humans.

One of the reactions I got to these remarks was: I don't know of any animal that can play Go! This was in reference to the AlphaGo system, 
which made news last year as having beaten the world champion for the game and is widely perceived as being a deep neural network, even by some AI researchers.

Indeed, we don't know of animals that can play a game as complex as Go. 
But AlphaGo is not a neural network since its architecture is based on a collection of AI techniques that have been in the works for at least
fifty years.\footnote{Oren Etzioni laid out this argument in a talk he gave at UCLA in March, 2016.} 
This includes the minimax technique for two-player games, stochastic search, learning from self play, the use of evaluation functions to cut off minimax search trees,
and reinforcement learning, in addition to neural networks. Therefore, while a Go player can be viewed as a function that maps a board configuration (input) to an 
action (output), the AlphaGo player was not built using the function-based approach (i.e., by learning a single function from input-output pairs), but 
only some of its components were built that way. The issue here is not only about assigning credit, but more about whether a competitive Go-function can be small enough 
to be represented and estimated under current data-gathering, storage and computational thresholds. It would be quite interesting if this was the case, 
but we don't know the answer to this yet (I should also note that the AlphaGo system is a great example of what one can achieve today by integrating 
model-based and function-based approaches).

\section{Pushing Thresholds}

One cannot of course preclude the possibility of constructing a competitive Go-function purely from data, or similarly complex functions, even though we are not there today given current thresholds.
But this begs the following question: If it is a matter of thresholds, and given current successes, why not focus all of our attention on moving these
thresholds further? While there is merit to this proposal, which seems to be currently adopted by key industries, it does face challenges
that stem from both academic and policy considerations. I will address academic considerations next while leaving policy considerations to a later section.

From an academic viewpoint, the history of AI tells us to be quite cautious as we have seen similar phenomena before. 
Those of us who have been around long enough can recall the era of expert systems in the 1980s. At that time, we discovered
ways to build functions using {\em rules,} which were devised through ``knowledge engineering'' sessions, as called then. The functions created through
this process, called {\em expert systems} or {\em knowledge-based systems,} were claimed to achieve performance that surpassed human experts in some cases, 
particularly in medical diagnosis.\footnote{\label{foonote:expert systems}One academic outcome of the expert system era was the construction of a 
dedicated Masters degree at Stanford University, called the ``Masters in AI,'' which was separate from the M.S. in computer science degree and had significantly
looser course requirements. It was a 2-year program, with the second year dedicated to building an expert system (I was a member of the very last class that 
graduated from this program before it was terminated). I personally recall that one of the justifications for this program was that classical
computer science techniques can be harmful to the thinking needed for effectively building expert systems!}
Terms such as ``knowledge is power'' were coined then that symbolized a jubilant state of affairs, resembling what ``deep learning'' has come  to symbolize today.
The period following this era became known as the {\em AI Winter,} as we could finally {\em delimit} the class of applications that yielded to such systems which
fell short of AI ambitions.

While the current derivative for progress on neural networks has been impressive, it has not been sustained long enough to allow sufficient visibility into 
the following consequential question. How effective will function-based approaches be when applied to new and broader applications than those already
targeted, and particularly ones that mandate more stringent measures of success?
This question has two parts. The first asks about the class of cognitive tasks whose corresponding functions are simple enough to allow compact representations
that can be evaluated efficiently (as in neural networks) and whose estimation is within the reach of current thresholds---or thresholds we expect 
to attain in, say, ten to twenty years? The second part of the question alludes to the fact that these functions are only approximations of cognitive tasks (i.e., they
don't always get it right). How suitable or acceptable will these approximations be when targeting cognitive tasks that mandate measures of success that are tighter than 
those required by the currently targeted applications?

\section{The Power of Success}

Before I comment on policy considerations, let me highlight a relevant phenomena that recurs in the history of science, with AI being no exception.
I will call this the {\em bullied-by-success} phenomena, in reference to the subduing of a community into mainly pursing what is currently successful, 
at the expense of pursuing enough what may be more successful or needed on the longer term.\footnote{One cannot here but bring into 
attention the long time it took neural networks to have the impact they are now having.}

Going back to history, some of the perspectives promoted during the expert systems era can be safely characterized 
today as having been scientifically absurd (see, for example, Footnote~\ref{foonote:expert systems}). Yet, due to the perceived success of expert systems then, these perspectives
had a dominating effect on the course of scientific dialogue and direction, leading to a bullied-by-success community.\footnote{A colleague couldn't but joke that the broad machine 
learning community is being bullied now by the success of its deep learning sub-community just as the broader AI community has been bullied by the success 
of its machine learning sub-community.}
I saw a similar phenomena during the transition from logic-based approaches to probability-based approaches for commonsense reasoning (late 1980s). 
Popular arguments then, like ``people don't reason probabilistically,'' which I believe carries merit, were completely silenced when probabilistic approaches started solving 
commonsense reasoning problems that have defied logical approaches for more than a decade.
The bullied-by-success community then made even far more reaching choices in this case as symbolic logic almost disappeared from the AI curricula. Departments
that were viewed as world centers for representing and reasoning with symbolic logic barely offered any logic courses as a result. Now we are paying the price. Not realizing
that probabilistic reasoning attributes numbers to Boolean propositions in the first place, and that logic was at the heart of probabilistic reasoning except in its most simple forms,
we now have come to the conclusion that we need to attribute probabilities to more complex Boolean propositions and even to first-order sentences. The resulting frameworks
are referred to as first-order probabilistic models or relational probabilistic models and there is a great need for skill in symbolic logic to advance these formalisms. The only
problem is that this skill has almost vanished from within the AI community, leading to treatments that appear naive compared to what early AI researchers may have 
been capable of accomplishing.\footnote{Early AI researchers left us with a wealth of results on first-order deduction (e.g., resolution and unification) but basically left 
untouched the subject of {\em first-order model counting,} which is at the heart of current needs for probabilistic reasoning and machine learning.}

The blame for this phenomena cannot be assigned to any particular party. It is natural for the successful to overjoy and sometimes also inflate success. It is
expected that industry will exploit success in ways that may redefine the employment market and influence the academic interests of graduate students. It is also
understandable that the rest of the academic community may play along to maintain its survival (win a grant, get a paper in, attract a student). While each of these
behaviors seem rational locally, their combination can sometimes be harmful to scientific inquiry and hence irrational globally. Beyond raising awareness about
this recurring phenomena, which can by itself go a long way, decision makers at the governmental and academic levels bear a particular responsibility towards mitigating its negative effects. 
Senior members of the academic community also bear the responsibility of putting current developments in historical perspective, to empower
junior researchers in pursuing their genuine academic interests instead of just yielding to current successes.\footnote{I made 
these remarks at a dinner table that included a young machine learning researcher. His reaction: ``I feel much better now.'' Apparently, this young researcher was subjected 
to this phenomena by support-vector-machine (SVM) researchers during his PhD work, when SVMs were at their peak and considered to be ``it'' at the time. Another young vision researcher---when 
pressed on whether deep learning can address our ambitions in vision research---told me, ``The reality is that you cannot publish a vision paper today in a top conference if it does not
contain a deep learning component, which is kind of depressing.''}

\section{Policy Considerations}

Let me now address the policy concern with regards to focusing all of our attention on functions instead of also models.
The biggest concern here stems from the lack of {\em interpretability} of learned functions, and the lack of {\em explainability} of
their results. If a medical diagnosis system recommended surgery for a patient, the patient and their physician would need to know why.
If a self driving car killed someone, we need to know why. If a face recognition system led to a mistaken arrest, we also need to know why.
If a voice command unintentionally shut down a power generation system, this needs to be explained as well. Answering {\em Why?} questions is central
to assigning blame and responsibility, which happen to lie at the heart of legal systems. Models can be used to answer such questions, but
functions cannot. This brings me to two relevant reactions that I received to my perspective, as the first  will help me explain this particular difference further. 

This was a question on why a function would not qualify as a model. Consider an engineered system that allows us to blow air into a balloon, which
then raises a lever that is positioned on top of the balloon. The input to this system is the amount of air we blow (\(X\)), while the output is the position of the lever (\(Y\)). 
We can learn a function that captures the behavior of this system by collecting \(X\)-\(Y\) pairs and then estimating the function \(Y = f(X)\). 
While this function may be all we need for certain applications, it would not qualify as a model as it does not capture the system mechanism.
Modeling the mechanism is essential for explanation (e.g., {\em why} is the change in the lever position not a linear function of the amount of air blown?)
and for causal reasoning more generally (e.g., {\em what if} the balloon is pinched?).

In his upcoming, ``The Book of Why,'' Judea Pearl explains further the differences between a (causal) model and a function, even though he does not use the term ``function'' explicitly.
In Chapter~1, he writes ``There is only one way a thinking entity (computer or human) can work out what would happen in multiple scenarios, 
including some that it has never experienced before. It must possess, consult, and manipulate a mental causal model of that reality.'' 
He then gives an example of a navigation system that is based on either (1) reasoning with a map (model) or (2) consulting a GPS system that only
gives a list of left-right turns for arriving at a destination (function). 
The rest of the discussion focuses on what can be done with the model but not the function. Pearl's argument particularly focuses on how a model can handle novel scenarios,
e.g., encountering road blocks that invalidate the function recommendations, while pointing to the combinatorial impossibility of encoding such contingencies in the function as it
must have a bounded size.\footnote{This is similar to responses given to Searle's Chinese Room Argument (1980), which comes down to having a function (of bounded size) that
emulates competence in speaking Chinese. Searle basically missed the (impossible) combinatoric assumption underlying his argument.} 

Interestingly, Chapter~1 in the ``The Book of Why'' starts with a discussion of evolution from forty thousand years ago, while arguing that the ability to 
answer ``What if?'' and ``Why?'' questions is  what made us ``uniquely human.''\footnote{Dan Roth alerted me to another perspective on functions versus models based on 
the distinction between discriminative versus generative learning. In particular, when learning a function (discriminatively), all one can usually do is 
estimate the value of its output variables. However, a learned (generative) model can be used to answer more queries through reasoning. Hector Geffner made a related
observation that functions are perhaps more suitable when the ``goals'' are fixed.}

\shrink{
Moreover, in the context of discussing evolution earlier in the chapter, he writes ``\ldots we 
could ask (and answer!) questions that no other animal can: {\em What if?} and {\em Why?} What if we attack the enemy from behind? Why was it a bad idea to light a fire in this forest? 
This is the cognitive leap that made us uniquely human.''
}

\shrink{
We have seen some ramifications of these policy concerns in two contexts recently. 
The effort led by the White House last year, under the label ``Preparing for the Future of AI,'' emphasized the {\em safety} of AI systems,
a property that can be hard to ensure without interpretability and explainability and thus models (the word ``safety'' appeared dozens of times  in the final report of this effort). 
DARPA is also starting a new program, called ``explainable AI'' or XAI, which is geared towards these considerations.
}

\section{A Theory of Cognitive Functions}

The second reaction I received to my model-based versus function-based perspective 
was during a workshop dedicated to deep learning at the Simons institute (March, 2017). The workshop title was actually ``representation
learning,'' a term that is being used with increased frequency by deep learning researchers. If you have followed presentations
on deep learning, you will notice that a critical component of getting these systems to work amounts to finding the correct architecture of the neural network.
Moreover, these architectures vary depending on the task and some of their components are sometimes portrayed as doing something that can be described
at an intuitive level. The reaction here was that deep learning is not learning a function (black box), but learning a representation (since the architecture is
not arbitrary but driven by the given task). I see this differently. Architecting the structure of a neural network is ``function engineering'' not ``representation learning,'' particularly
since the structure is penalized and rewarded by virtue of its conformity with input-output pairs.
The outcome of function engineering amounts to restricting the class of functions that can be learned using parameter estimation techniques.
This is akin to restricting the class of distributions that can be learned after one fixes the topology of a probabilistic graphical model.
The practice of representation learning is then an exercise of identifying the classes of functions that are suitable for certain tasks.

In this connection, I think what is needed most is a theory of {\em cognitive functions.}
A cognitive function is one that captures a relationship that is typically associated with cognition, such as mapping audio signals to words, or mapping words
to some meaning. What seems to be needed is a catalogue of cognitive functions and a study of their representational complexity (the size and 
nature of architectures needed to represent them) in addition to a study of their learnability and approximateability. 
For Boolean functions, we have a deep theory of this kind. In particular, researchers have cataloged 
various functions in terms of the space needed to represent them in different forms (e.g., CNFs, DNFs, OBDDs). What we need is something similar
for real-valued functions that are meant to capture cognitive behaviors. In a sense, we already have some leads into such a theory---for example,
researchers seem to know what architectures (aka, function classes) can be more effective for certain object identification tasks. This needs to be formalized
and put on solid theoretical ground. This theory would also include results on the learnability of function classes using estimation techniques
employed by the deep learning community (gradient descent in particular). Interestingly, such results were presented at the representation learning workshop referenced above
(talk titled ``failures of deep learning''), where very simple functions were presented which defeat current estimation techniques. Even more interestingly,
some have dismissed the importance of such results in side discussions on the grounds that the identified functions are not of practical
significance (read ``these are not cognitive functions'' or ``we've gotten a long way by learning approximations to functions''). In fact, if I had my say, I would rename 
the field of deep learning to the field of {\em learning approximations of cognitive functions.}

\section{Concluding Reflections}

This article was motivated by concerns that I and others have on how current progress in AI is being framed and perceived today. 
Without a scholarly discussion of the causes and effects of recent achievements, and without a proper 
perspective on the obtained results, one stands to hinder further progress by
perhaps misguiding the young generation of researchers or misallocating resources at the academic, industrial and governmental levels. 
One also stands to misinform a public that has developed a great interest in artificial intelligence and its implications. 
The current negative discussions by the public on the AI singularity, also called super intelligence, can only be attributed to the lack of
accurate framings and characterizations of recent progress. With almost everyone being either overexcited by the new developments or overwhelmed by them,
substantial scholarly discussions and reflections went missing. 

I had the privilege of starting my research career in AI around the mid to late 1980s, during one of the major crises of the field (a period
marked by inability instead of ability). I was dismayed then as I sat in classes at Stanford University, witnessing how AI researchers were being significantly
challenged by some of the simpler tasks performed routinely by humans.
In retrospect, I now realize how such crises can be enabling for scientific discovery as they fuel academic thinking, empower researchers, and create grounds for
profound scientific contributions.\footnote{The seminal work of Judea Pearl on probabilistic approaches to commonsense reasoning is
one example outcome of the mentioned crisis.} On the other hand, I am reminded today how times of achievements can potentially slow down scientific progress
by shifting academic interests, resources and brain power too significantly towards exploiting what was just discovered, at the expense of understanding
such discoveries, and preparing for the moment when their practical applications have been delimited or exhausted. There are many dimensions for this
preparation. For the deep learning community, perhaps the most significant one being a transition from the ``look what else {\em we} can do'' mode 
into a ``look what else {\em you} can do'' mode. This is not only an invitation to reach out to and empower the broader AI community. 
It is also a challenge since such a transition is not only a function of attitude but also an ability to characterize progress in ways that 
enable people from outside the community to understand, characterize and capitalize on. The broader AI community bears no less responsibility
for this preparation. This community is also both invited and challenged to identify fundamental ways in which functions can be tuned into a boon 
for building and learning models, and to provide contexts that facilitate this identification by AI researchers across the spectrum.

I will now conclude with this additional reflection. I wrote the first draft of this article in November, 2016. A number of colleagues provided positive 
feedback then, with one warning me about a negative tone that comes out of the text. I put the draft on hold for a few months as a result, 
while continuing to share its contents verbally in various contexts and revising it accordingly. The decision to release it now was triggered by two events:
a discussion of these thoughts at a workshop  organized by the law school at UCLA, and other discussions with colleagues outside of AI (architecture, programming languages, networks and theory).
These discussions revealed the substantial interest in this subject and led me to conclude that the most important objective I should be seeking is ``starting a discussion.''
I may have erred in certain parts, I may have failed to give due credit, and I may have missed parts of the evolving scene. 
I just hope that the thoughts I am sharing here will start a discussion and that the collective wisdom of the community will correct what I may have gotten wrong.

\section*{Acknowledgements}
I benefited greatly from discussions and feedback that I received from colleagues who are too many to enumerate here, but whose input was critical
to shaping the thoughts expressed in this article.

\end{document}